\documentclass[10pt,twocolumn,letterpaper]{article}

\usepackage{wacv}
\usepackage{times}
\usepackage{epsfig}
\usepackage{graphicx}
\usepackage{amsmath}
\usepackage{amssymb}

\usepackage{multirow}
\usepackage{enumitem}
\usepackage{subcaption}
\usepackage{color,soul}
\newcommand{\norm}[1]{\left\lVert#1\right\rVert}

\def\wacvPaperID{20} 

\wacvfinalcopy 

\ifwacvfinal
\def\assignedStartPage{1} 
\fi

\ifwacvfinal
\usepackage[breaklinks=true,bookmarks=false]{hyperref}
\else
\usepackage[pagebackref=true,breaklinks=true,colorlinks,bookmarks=false]{hyperref}
\fi

\ifwacvfinal
\else
\fi

\begin{document}

\title{Towards Fair Cross-Domain Adaptation via Generative Learning}

\author{Tongxin Wang\textsuperscript{\rm 1}\qquad Zhengming Ding\textsuperscript{\rm 2, \footnotemark[1]}\qquad Wei Shao\textsuperscript{\rm 3}\qquad Haixu Tang\textsuperscript{\rm 1}\qquad Kun Huang\textsuperscript{\rm 3, 4, }\thanks{Corresponding authors.}\\
		{\textsuperscript{\rm 1} Department of Computer Science, Indiana University Bloomington}
		{\textsuperscript{\rm 2} Department of CIT, IUPUI}\\
		{\textsuperscript{\rm 3} Department of Medicine, IU School of Medicine}
		{\textsuperscript{\rm 4} Regenstrief Institute} \\
		\small{\texttt{\{tw11,zd2,shaowei,kunhuang\}@iu.edu}} \qquad
		\small{\texttt{hatang@indiana.edu}}
	}

\maketitle

\begin{abstract}
Domain Adaptation (DA) targets at adapting a model trained over the well-labeled source domain to the unlabeled target domain lying in different distributions. Existing DA normally assumes the well-labeled source domain is class-wise balanced, which means the size per source class is relatively similar. However, in real-world applications, labeled samples for some categories in the source domain could be extremely few due to the difficulty of data collection and annotation, which leads to decreasing performance over target domain on those few-shot categories. To perform fair cross-domain adaptation and boost the performance on these minority categories, we develop a novel Generative Few-shot Cross-domain Adaptation (GFCA) algorithm for fair cross-domain classification. Specifically, generative feature augmentation is explored to synthesize effective training data for few-shot source classes, while effective cross-domain alignment aims to adapt knowledge from source to facilitate the target learning. Experimental results on two large cross-domain visual datasets demonstrate the effectiveness of our proposed method on improving both few-shot and overall classification accuracy comparing with the state-of-the-art DA approaches.
\end{abstract}

\vspace{-5mm}
\section{Introduction}
\vspace{-2mm}
In recent years, deep learning has achieved significant advances in various applications. In most real-world applications, the availability of large-scale labeled data is crucial. However, manually collecting and annotating data are extremely expensive and time consuming for every new domain. On the other hand, we can often access abundant data with limited or even no labels. \textit{Domain adaptation} (DA) \cite{pan2009survey,wang2018deep} considers the problem of transferring a machine learning model from a source domain towards a different target domain, where data from different domains lie in different distributions. By reducing both the marginal and conditional mismatch in feature space between domains, knowledge transfer from an external labeled, well-established source domain data to target domain can be achieved.
Existing DA methods generally deal with domain adaptation by assuming source classes are balanced. 
However, in reality, labeled data from some categories may be limited due to the difficulty of data collection and annotation, making data in source domain extremely imbalanced.
We refer to this as \textit{Few-shot Cross-domain Adaptation}, where the classes with limited samples denote \textit{few-shot classes} while others represent normal classes.
The few-shot cross-domain adaptation problem raises concern when considering the fairness of the corresponding learning task as the available training data is already contaminated by bias towards classes with majority of samples \cite{chouldechova2018frontiers}.
Existing DA methods can be roughly summarized into two categories: discrepancy-based methods and adversarial-based methods \cite{wang2018deep}. Discrepancy-based methods focus on diminishing the domain shift through minimizing discrepancy metrics, such as maximum mean discrepancy (MMD) \cite{long2015learning,tzeng2014deep} and correlation alignment (CORAL) loss \cite{sun2016return,sun2016deep,zhuo2017deep}. 
More recently, inspired by the idea of generative adversarial networks \cite{goodfellow2014generative}, adversarial-based methods aiming to match feature distributions through adversarial training have attracted great attentions \cite{ganin2014unsupervised,liu2016coupled,tzeng2017adversarial}.
Methods in both categories are designed to reduce the cross-domain gap by aligning domain-wise distributions explicitly, so that the model derived from the source domain can be applied to the target domain directly.

Dealing with imbalanced data is commonly encountered in real-world applications where some classes may have very limited labeled training data. 
However, directly minimizing the average error to fit the entire training data could make learning techniques bias towards the majority population and lead to higher distribution of errors and unfair results in the minority population \cite{chouldechova2018frontiers}.
Moreover, the idea of learning novel concepts from very few examples has attracted much attention in few-shot learning approaches \cite{dvornik2019diversity,peng2019few,tokmakov2019learning,zhang2019variational}.
Among the existing approaches, data augmentation is a straightforward method to boost the performance on few-shot classes through synthesizing more data in input space  \cite{hariharan2017low,schwartz2018delta,wang2018low,zhang2018metagan} or in a learned feature space \cite{devries2017dataset,volpi2018adversarial}. 
Generative models have been utilized to synthesize more training data for few-shot classes through capturing the data variation from base classes \cite{goodfellow2014generative,mirza2014conditional}. Moreover, extended from typical DA problem, methods have been proposed to directly tackle the problem of imbalanced data in DA \cite{ando2017deep,ming2015unsupervised,wang2017balanced}. Existing DA methods focus on sample re-weighting \cite{ando2017deep}, instance re-weighting with subspace learning \cite{ming2015unsupervised}, and class re-weighting with distribution adaptation \cite{wang2017balanced}. However, these methods did not fully utilize the powerful generative models to directly augment samples for underrepresented few-shot classes.

To this end, we propose a novel Generative Few-shot Cross-domain Adaptation model (GFCA) to enhance the adaptation ability from extremely imbalanced source domain for better and fairer target sample prediction. Specifically, we focus on generating effective auxiliary data for the few-shot classes in the source domain to facilitate the classification accuracy for the under-represented classes. To the best of our knowledge, this is the very pioneering work to explore the few-shot cross-domain adaptation problem for fair cross-domain classification.
The main contributions of our paper are summarized as follows:
\begin{itemize}
\vspace{-2mm}
  \item We incorporate the generative adversarial network in training a DA network for both normal and few-shot classes. Specifically, the generator attempts to synthesize effective fake data at the learned feature space to augment few-shot classes, while the discriminator guides the data generation in order to adapt data variation in normal classes to generate data in few-shot classes.
  \vspace{-2mm}
  \item We combine the generative adversarial network with the DA network to train a general classifier for both source and target domain. Specifically, we build a conditional generative adversarial network to promote the performance of the DA network on few-shot classes.
  \vspace{-2mm}
  \item We evaluate our proposed model on two cross-domain visual benchmarks. Comparing with state-of-the-art methods, our model achieves significant improvement in both few-shot and overall classification accuracy, which indicates fairer classification capability.
\end{itemize}

\vspace{-4mm}
\section{Related Work}
\vspace{-2mm}
\begin{figure*}[t]
\begin{center}
\includegraphics[width=0.8\linewidth]{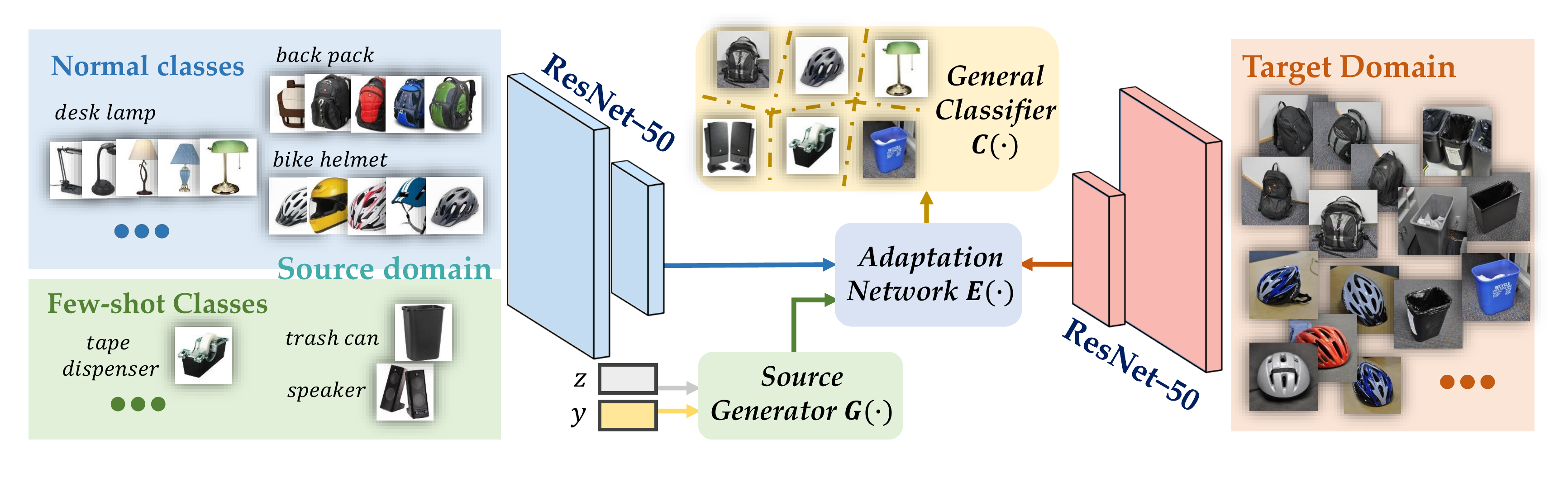}\vspace{-4mm}
\caption{
Illustration of the few-shot cross-domain adaptation problem with GFCA. Source domain contains labeled data of normal and few-shot classes without overlapping classes. Target domain contains unlabeled data sharing the same set of classes as source domain. Representation learning trains effective ResNet-50 feature extractor using the labeled source data to transfer the image into a feature space for both source and target domains. Generative data augmentation is used to synthesize effective training samples from random noise vector $z$ and one-hot label vector $y$ by training source generator $G(\cdot)$. Cross-domain adaptation seeks a general classifier $C(\cdot)$ for both source and target domain through effective cross-domain alignment by training the adaptation network $E(\cdot)$.}\label{fig:method}
\end{center}
\vspace{-10mm}
\end{figure*}

In this section, since few-shot cross-domain adaptation is still an open problem, we mainly review two research areas that are related to the proposed model: data augmentation and unsupervised DA.

\vspace{-2mm}
\subsection{Data Augmentation}
\vspace{-1mm}
Data augmentation is a straightforward approach to improve the performance on few-shot classification. 
Generating data with enough intra-class variation in few-shot classes is crucial to effective data augmentation. 
Hariharan et al. \cite{hariharan2017low} proposed techniques to "hallucinate" additional training examples for few-shot classes by leveraging the mode of variation in the normal classes. 
Hauberg et al. \cite{hauberg2016dreaming} proposed to learn augmentation strategy on a per-class basis, where a statistical model is built for transformations within a given class and used for augmenting the dataset.
DeVries and Taylor \cite{devries2017dataset} proposed to augment data in the learned feature space using simple transformations.
Recently, deep generative models \cite{antoniou2017data,goodfellow2014generative,mariani2018bagan,mirza2014conditional,odena2017conditional} have been exploited to synthesize data for few-shot classes by capturing the intra-class variation in normal classes. 
Data Augmentation GAN (DAGAN) \cite{antoniou2017data} synthesizes new data by adding noise to lower-dimensional representations of the data learned through a generator network with an encoder and a decoder. 
BAlancing GAN (BAGAN) \cite{mariani2018bagan} also utilizes an autoencoder for data generation and learns useful features from normal classes for few-shot data synthesis by applying class conditioning on the latent space in the autoencoder. 
However, the key difference between GFCA and existing works is that we utilize data generation in the feature domain rather than in the image domain using deep generative models while simultaneously seeking DA between source and target domains. 

\vspace{-2mm}
\subsection{Unsupervised Domain Adaptation}
\vspace{-1mm}
Unsupervised DA aims to bridge the distribution difference between domains with unlabeled target domain data. Recently, deep DA methods have been attracting popularity by utilizing the power of deep neural networks.
Discrepancy-based methods aim to diminish the domain shift by minimizing certain discrepancy metrics \cite{lee2019sliced,long2015learning,sun2016deep,tzeng2014deep,zhuo2017deep}.  
Deep Domain Confusion (DDC) network \cite{tzeng2014deep} incorporates an additional loss based on the MMD between the source and target representations at the last fully-connected layer to jointly optimize for classification and domain invariance. Extended from DDC, Domain Adaptation Network (DAN) \cite{long2015learning} utilizes multiple kernel MMD-based loss across domains on the last three task-specific layers and achieves better performance. 
More recently, Lee et al. \cite{lee2019sliced} proposed Sliced Wasserstein Discrepancy (SWD), which can be minimized between task-specific classifiers to align the distributions across domains. 
Saito et al. \cite{saito2018maximum} presented MCD-DA to align distributions across domains by exploiting task-specific decision boundaries. 
Xu et al. \cite{xu2019larger} introduced to adapt the feature norms with larger-norm constraint for more informative and transferable computation across domains.

Inspired by generative adversarial networks \cite{goodfellow2014generative}, adversarial-based methods utilize adversarial training to align feature distributions across domains. Domain-Adversarial Neural Network (DANN) \cite{ganin2014unsupervised} is one of the first methods to learn domain-invariant representations through adversarial learning between the domain classification and feature representation. Tzeng et al. \cite{tzeng2017adversarial} 
proposed a unified framework for unsupervised DA and 
further introduced Adversarial Discriminative Domain Adaptation (ADDA) by minimizing a GAN-based loss. 
Volpi et al. \cite{volpi2018adversarial} combined feature augmentation with domain-invariance enforcement for more effective domain adaptation, but did not include mechanism to promote the underrepresented classes in the training set.
More recently, Cao et al. \cite{cao2018partial} presented Partial Adversarial Domain Adaptation (PADA)
for more effective adaptation when source domain and target domain do not share identical label space. 
Long et al. \cite{long2018conditional} proposed Conditional Domain Adversarial Network (CDAN) to utilized the discriminative information in classifier predictions to help adversarial adaptation.
Zhang et al. \cite{zhang2019bridging} introduced the well-designed theory-induced Margin Disparity Discrepancy (MDD) for adversarial training in DA. In this paper, to demonstrate the necessity of leveraging generative models for data augmentation, we utilize the well-established DAN method for DA based on the real source data, synthetic source data, and target data.

\vspace{-2mm}
\section{The Proposed Algorithm}
\vspace{-1mm}
\subsection{Preliminary and Motivation}
\vspace{-2mm}
In the unsupervised DA problem, we are given a labeled source domain and an unlabeled target domain. The source domain contains $n_s$ data points from $c$ classes: $\{X_s, Y_s\} = \{(x_{s,1}, y_{s,1}), ..., (x_{s,n_s}, y_{s,n_s})\}$, where $x_{s,i} \in \mathbb{R}^{d_x}$ is the feature vector with $d_x$ dimensions and $y_{s,i} \in \mathbb{R}^c$ is the corresponding one-hot label vector. The target domain contains $n_t$ data points: $\{X_t\} = \{x_{t,1}, ..., x_{t,n_t}\}$, in which $x_{t,i} \in \mathbb{R}^{d_x}$. Source and target domains are different in terms of data distributions, but share consistent label information, and the goal is to recognize the unlabeled target samples.

In Few-shot Cross-domain Adaptation problem, we further assume that the labeled source domain data contains two sets of data without overlap in classes: \textbf{Normal Set} (i.e., normal classes) $\{X_{sn}, Y_{sn}\}$ with $c_n$ classes and \textbf{Few-Shot Set} (i.e., few-shot classes) $\{X_{sl}, Y_{sl}\}$ with $c_l$ classes, where $\{X_s, Y_s\} = \{X_{sn}, Y_{sn}\} \cup \{X_{sl}, Y_{sl}\}$ and $c = c_n + c_l$. In this problem, few-shot set contains far less number of samples per class, and our goal is to achieve high classification performance in both normal classes and few-shot classes on the unlabeled target samples.

In reality, labeled data from some classes may be limited due to the difficulty of data collection and annotation. While existing DA methods generally assume source classes to be balanced, few-shot cross-domain adaptation is a challenging task because of extremely imbalanced source domain. Few-shot classes only contains limited labeled training data with limited variation within each class. Traditional data augmentation methods \cite{hariharan2017low} use human designed rules to generate more data for the few-shot classes, where the improvement to the classifier is limited. Moreover, it is important to balance between the normal set and the few-shot set, since trying to improve the performance on the few-shot classes could hurt normal set classification. Thus, it is important to generate effective data to expand the intra-class variation in few-shot classes. Generative models have shown great promise in synthesizing effective data. Since normal set usually contains larger intra-class variations, generative models can be leveraged to adapt the variation in normal set to few-shot set during data generation. Once we properly augment the source domain data to promote few-shot classes, we can adopt DA approaches to train a general classifier for both source and target domain and both normal and few-shot sets. In this paper, we utilize MMD-based regularization for cross-domain alignment, where a classifier can be subsequently trained to effectively classify data from both domains.

\vspace{-3mm}
\subsection{Generative Few-Shot Cross-Domain Learning}
\vspace{-2mm}

An illustration of our proposed Generative Few-shot Cross-domain Adaptation (GFCA) algorithm is shown in Figure \ref{fig:method}. An representation model is built first with supervised learning framework to transfer the image into a discriminative feature domain by training on the source domain data. Taking the input of a random noise and a one-hot label vector, the generator is trained to synthesize effective source domain data with the guidance of the discriminator in the learned feature space. A MMD-regularized encoder and classifier is then trained using real source domain data, synthetic fake source domain data, and target domain data to perform effective classification on data from both domains.

\vspace{-5mm}
\subsubsection{Generative Data Augmentation}
\vspace{-2mm}
Generative models trained via adversarial training can synthesize fake data from random noise \cite{goodfellow2014generative}. Moreover, conditional generative models have been proved to be more effective in synthesizing meaningful data by conditioning the model on additional information such as class labels \cite{mirza2014conditional}. For generative data augmentation, our goal is to use generative models to generate fake data at the learned feature space with class labels for augmenting source domain training data. We denote the random noise by $z \in \mathbb{R}^{d_z}$, real feature vector by $x \in \mathbb{R}^{d_x}$, and the corresponding one-hot label vector by $y \in \mathbb{R}^c$. 

The generator $G(\cdot)$ attempts to synthesize fake feature vector $x_f$ with the inputs of $z$ and $y$:
\vspace{-2mm}
\begin{equation}
\begin{split}
G(z|y) &= \phi_g(W_g\begin{bmatrix} z\\ y \end{bmatrix}) = \phi_g(\begin{bmatrix} W_z, W_y \end{bmatrix} \begin{bmatrix} z\\ y \end{bmatrix}) \\
&= \phi_g(W_z z + W_y y),
\end{split}
\vspace{-2mm}
\end{equation}
where $W_z \in \mathbb{R}^{d_x \times d_z}$ and $W_y \in \mathbb{R}^{d_x \times c}$ are the weight matrices. $\phi_g(\cdot)$ is the element-wise activation function.

We attempt to augment few-shot classes in the source domain by synthesizing fake data using generator $G(\cdot)$ in the learned feature space. To acquire meaningful synthetic features, we could span the feature space of each few-shot class around its center. Thus, we use $W_y y$ to capture the class center information and use $W_z z$ to capture the class variance information. 
We initialize $W_y$ with the class centroid of source domain features in the training set. 
The normal set contains abundant samples for each class, which makes the estimation of class centers more accurate than class centers of few-shot set during initialization. Therefore, we include an additional L2 regularization on the weight vectors for the normal set in $W_y$ to discourage it from changing dramatically comparing to initialization.
For the random noise part, to faithfully capture the class variation, we initialize each column of $W_z$ by taking the top $d_z$ principle components from the source training data $X_s$ and scale each column of $W_z$ by multiplying the corresponding eigenvalues. 
Each element in $z$ is sampled from a uniform distribution between -1 and 1. We also add a normalization process to make the synthetic feature vectors be in the same scale as the real source features. For a fake feature vector $x_f$ generated by $G(\cdot)$, we normalize $x_f$ as $N(x_f) = x_f / \norm{x_f}_2 * \beta$, where $\beta$ is the average norm of real source domain features. 

The discriminator $D(\cdot)$ attempts to distinguish the fake feature vectors from the real ones, which is designed as:
\vspace{-2mm}
\begin{equation}
D(x) = \phi_d(W_d x + b_d),
\vspace{-2mm}
\end{equation}
where $W_d \in \mathbb{R}^{1 \times d_x}$ and $b_d \in \mathbb{R}$. $\phi_d(\cdot)$ projects the input to a value between 0 and 1.

The generator $G(\cdot)$ aims to generate features similar to real source features, while the discriminator $D(\cdot)$ aims to differentiate fake features from real features. Therefore, the loss function for the generator $L_g$ and for the discriminator $L_d$ can be formulated as:
\begin{equation}
L_g = -\mathbb{E}[D(x_f)]
\end{equation}
\begin{equation}
L_d = -\mathbb{E}[D(x_r)] - \mathbb{E}[1-D(x_f)]
\end{equation}
where $x_r$ represents the real feature vector and $x_f = G(z|y)$ represents the fake feature vector given one-hot label $y$ and random noise $z$. As shown above, the generator is trained by minimizing $L_g$, while the discriminator is trained by minimizing $L_d$. Thus, the training of generator $G(\cdot)$ and discriminator $D(\cdot)$ can be formulated as a two-player minimax problem.

\vspace{-3mm}
\subsubsection{Cross-Domain Alignment and Fair Classification}
\vspace{-2mm}
Maximum Mean Discrepancy (MMD) \cite{sejdinovic2013equivalence} is a non-parametric distance estimate between two probability distributions based on their samples. MMD-based regularization has proven to be highly effective in many DA tasks \cite{long2015learning,long2017deep,sejdinovic2013equivalence}. Therefore, we utilize MMD to bridge the distribution gap between the learned representations of source and target domain features. Given samples $X_s = \{x_{s,i}\}_{i = 1,...,n_s}$ drawn from distribution $\mathcal{D}_s$ and $X_t = \{x_{t,i}\}_{i = 1,...,n_t}$ drawn from distribution $\mathcal{D}_t$, the empirical estimate of MMD can be written as
\begin{equation}
MMD(X_s, X_t) = \norm{\frac{1}{n_s}\sum_{i=1}^{n_s}\phi(x_{s,i}) - \frac{1}{n_t}\sum_{j=1}^{n_t}\phi(x_{t,j})}_{\mathcal{H}}
\end{equation}
where $\phi(\cdot)$ is the feature space map from the original feature space to a universal Reproducing Kernel Hilbert Space (RKHS) $\mathcal{H}$. Specifically, in this paper, we utilize a multiple kernel variant of MMD (MK-MMD) proposed by Gretton et al. \cite{gretton2012optimal}, which is also the backbone of DAN \cite{long2015learning}.

We exploit an encoder $E(\cdot)$ for learning transferable features with dimension $d_h$ between source and target domains. For samples from the source domain $X_s$ and the target domain $X_t$, the encoder attempts to minimize $L_e$, which is defined as the MMD between their corresponding representations $H_s$ and $H_t$ produced by the encoder:
\begin{equation}
L_e = MMD(H_s, H_t) = MMD(E(X_s), E(X_t))
\end{equation}

Finally, we train a classifier $C(\cdot)$ with $c$ classes and weight matrix $W_c \in \mathbb{R}^{d_c \times d_h}$ to classify samples from both domains. Since there are only limited labeled training samples for the $c_l$ few shot classes, in order to obtain a good classifier for few-shot classification, we use the generator to synthesize additional labeled samples from the source domain for training. Therefore, the classification loss can be formulated as
\vspace{-2mm}
\begin{equation}
\begin{split}
L_c =& L_{sr}+L_{sf} \\
=& -\mathbb{E}[\log P(Y=y_r|E(x_{sr}))] \\ 
& -\mathbb{E}[\log P(Y=y_f|E(G(z|y_f)))]
\end{split}
\end{equation}
where $L_{sr}$ is the classification loss for labeled real data from source domain $x_{sr}$ with label $y_r$ and $L_{sf}$ is the classification loss for fake source domain feature vector $x_{sf}$ generated by $G(\cdot)$ with label $y_f$.

To further balance between the few-shot set with limited training samples and the normal set, we incorporate a weight regularizer on the classifier using a fair classification (FC) term \cite{guo2017one} for fairer classification across few-shot and normal classes. Denoting the classifier weight vector for the $k$-th class by $w_k$, the FC term can be formulated as
\vspace{-2mm}
\begin{equation} \label{eq:fc}
L_{fc} = (\frac{1}{C_l} \sum\nolimits_{k \in C_l} \norm{w_k}_2^2 - \alpha)^2
\vspace{-2mm}
\end{equation}
where $C_l$ is the set of the class indices for the few-shot set. $\alpha$ is the parameter learned from the average of the squared norms of weight vectors for the normal classes:
\vspace{-2mm}
\begin{equation}
\alpha = \frac{1}{|C_n|}\sum\nolimits_{k \in C_n}\norm{w_k}_2^2
\vspace{-2mm}
\end{equation}
where $C_n$ is the set of the class indices for the normal set. 
The FC term is based on the assumption that on average, samples in the few-shot classes should occupy a space of similar volume in the feature space as the normal classes \cite{guo2017one}. Therefore, in Eq. \ref{eq:fc}, the average of the squared norms of the weight vectors for the few-shot classes are regularized to the same scale as normal classes.

In summary, the overall objective for cross domain alignment and few-shot promotion can be written as
\vspace{-3mm}
\begin{equation}\label{eq:ec}
L_{ec} = L_c + \lambda L_e + \gamma L_{fc}
\vspace{-3mm}
\end{equation}
where $\lambda$ and $\gamma$ controls the relative importance of the MMD regularization term and the fair classification term. The encoder $E(\cdot)$ and classifier $C(\cdot)$ are trained jointly by minimizing $L_{ec}$.

\vspace{-4mm}
\subsubsection{Implementation Details}
\vspace{-2mm}
We choose the standard residual network with 50 layers (ResNet-50) \cite{he2016deep} pre-trained on the ImageNet dataset as the backbone of our feature extractor. We fine-tune the ResNet-50 network using the labeled source domain data for each task and use the last pooling layer as the image representation. The hyper-parameters $\lambda$ and $\gamma$ in the Eq. \ref{eq:ec} are selected as 1 throughout all experiments. We use the sigmoid activation function for $D(\cdot)$, and the leaky-ReLU activation function for $G(\cdot)$, $E(\cdot)$, and $C(\cdot)$. We adopt the adaptive moment
estimation (Adam) to train the network with $\beta_1=0.9$ and $\beta_2=0.999$. The generator $G(\cdot)$ and discriminator $D(\cdot)$ is first pre-trained separately to acquire a good initialization of the generative feature augmentation network. During the end-to-end training of the entire network, for each iteration, we first sample real labeled source domain data and unlabeled target domain data. Fake labeled source domain features are then synthesized through the generator. The encoder and classifier are then updated to minimize $L_{ec}$. Next, we fix the discriminator and optimize the generator by minimizing $L_g$. Finally, we constrain the generator to update the discriminator by minimizing $L_d$. Thus, we alternatively update different modules within GFCA until it converges.

\vspace{-2mm}
\section{Experiments}

\begin{table*}[ht]
\centering
\small
\caption{Accuracy (\%) of few-shot classification for DA tasks in the Office31 dataset, where A = Amazon, D = DSLR, and W = Webcam. Avg\_l, Avg\_n, and Avg denote the average classification accuracy for few-shot classes, normal classes, and all unlabeled target samples, respectively.}\vspace{-3mm}
\begin{tabular}{lcccccc|ccc}
\hline
Methods & A$\rightarrow$W & D$\rightarrow$W & W$\rightarrow$D & A$\rightarrow$D & D$\rightarrow$A & W$\rightarrow$A & Avg\_l & Avg\_n & Avg\\
\hline
\textbf{ResNet-50} & 19.3$\pm$0.2 & 57.9$\pm$0.5 & 51.2$\pm$0.3 & 22.6$\pm$0.3 & 23.8$\pm$0.7 & 24.4$\pm$0.3 & 33.2 & 73.5 & 60.1  \\
\textbf{DAN} & 63.5$\pm$0.1 & 71.7$\pm$0.4 & 86.6$\pm$0.1 & 67.4$\pm$0.2 & 47.9$\pm$0.1 & 51.3$\pm$0.1 & 64.7 & 84.3 & 77.9  \\
\textbf{DANN} & 37.8$\pm$0.2 & 79.8$\pm$0.1 & 85.4$\pm$0.2 & 44.9$\pm$0.4 & 38.6$\pm$0.0 & 41.1$\pm$0.1 & 54.6 & 86.0 & 75.6  \\
\textbf{ADDA} & 35.8$\pm$0.1 & 65.5$\pm$0.2 & 79.9$\pm$0.1 & 37.7$\pm$0.3 & 37.6$\pm$0.1 & 40.0$\pm$0.0 & 49.4 & 85.6 & 73.7  \\
\textbf{PADA} & 19.7$\pm$0.0 & 65.0$\pm$0.3 & 74.4$\pm$0.2 & 26.3$\pm$0.4 & 33.4$\pm$0.2 & 33.5$\pm$0.2 & 42.0 & 83.8 & 70.0  \\
\textbf{MDD} & 32.0$\pm$0.3 & 75.5$\pm$0.2 & 83.1$\pm$0.3 & 39.0$\pm$0.3 & 35.3$\pm$0.3 & 39.1$\pm$0.2 & 50.6 & 86.5 & 74.6  \\
\textbf{SWD} & 25.5$\pm$0.1 & 70.0$\pm$0.2 & 77.6$\pm$0.0 & 34.4$\pm$0.2 & 34.1$\pm$0.1 & 36.4$\pm$0.1 & 46.3 & 83.6 & 71.3  \\
\textbf{CDAN} & 35.1$\pm$0.1 & 71.8$\pm$0.1 & 76.2$\pm$0.2 & 42.9$\pm$0.5 & 33.8$\pm$0.1 & 34.8$\pm$0.1 & 49.1 & 86.0 & 73.8 \\
\textbf{MCD-DA} & 33.5$\pm$0.1 & 71.6$\pm$0.0 & 76.1$\pm$0.1 & 41.6$\pm$0.2 & 37.2$\pm$0.1 & 39.1$\pm$0.1 & 49.9 & 84.0 & 72.7 \\
\textbf{DIFA} & 63.0$\pm$0.4 & 67.2$\pm$0.2 & 89.2$\pm$0.1 & 58.3$\pm$0.5 & 49.7$\pm$1.0 & 50.7$\pm$0.1 & 63.0 & 84.5 & 77.6 \\
\textbf{SAFN} & 40.3$\pm$0.4 & 72.0$\pm$0.1 & 77.9$\pm$0.2 & 48.8$\pm$0.2 & 37.8$\pm$0.0 & 38.2$\pm$0.0 & 52.5 & 86.0 & 74.9 \\
\hline
\textbf{GFCA-2Stage} & 71.3$\pm$0.0 & \textbf{81.2$\pm$0.1} & 90.5$\pm$0.2 & 73.4$\pm$0.6 & 46.3$\pm$0.0 & 50.3$\pm$0.0 & 68.8 & 86.6 & 80.8  \\
\textbf{GFCA-WoFC} & \textbf{71.6$\pm$0.1} & 80.1$\pm$0.2 & \textbf{90.7$\pm$0.2} & 74.6$\pm$0.9 & \textbf{53.0$\pm$0.0} & 55.7$\pm$0.0 & \textbf{71.0} & 86.7 & 81.6  \\
\textbf{GFCA} & 71.3$\pm$0.1 & 80.3$\pm$0.2 & 90.5$\pm$0.2 & \textbf{74.9$\pm$0.8} & 52.8$\pm$0.0 & \textbf{55.8$\pm$0.0} & 70.9 & \textbf{86.9} & \textbf{81.7}  \\
\hline
\end{tabular}
\vspace{-3mm}
\label{tbl:office31}
\end{table*}

\begin{table*}[ht]
\centering
\small
\caption{Accuracy (\%) of few-shot classification for DA tasks in the Office-Home dataset, where Ar = Art, Cl = Clipart, Pr = Product, and Rw = Real-world. Avg\_l, Avg\_n, and Avg denote the average accuracy for few-shot classes, normal classes, and all unlabeled target samples, respectively.}\vspace{-3mm}
\begin{tabular}{lcccccccccccc|ccc}
\hline
\vspace{-2mm}
\multirow{ 3}{*}{Methods} & {Ar} & {Ar} & {Ar} & {Cl} & {Cl} & {Cl} & {Pr} & {Pr} & {Pr} & {Rw} & {Rw} & {Rw} & \multirow{ 3}{*}{\text{Avg\_l}} & \multirow{ 3}{*}{\text{Avg\_n}} & \multirow{ 3}{*}{\text{Avg}}\\
\vspace{-1mm}
& {\tiny$\downarrow$} & {\tiny$\downarrow$} & {\tiny$\downarrow$} & {\tiny$\downarrow$} & {\tiny$\downarrow$} & {\tiny$\downarrow$} & {\tiny$\downarrow$} & {\tiny$\downarrow$} & {\tiny$\downarrow$} & {\tiny$\downarrow$} & {\tiny$\downarrow$} & {\tiny$\downarrow$} &  &  & \\

& {Cl} & {Pr} & {Rw} & {Ar} & {Pr} & {Rw} & {Ar} & {Cl} & {Rw} & {Ar} & {Cl} & {Pr} &  &  & \\

\hline
\textbf{ResNet-50} & 11.9 & 35.0 & 21.1 & 14.7 & 24.2 & 23.7 & 10.5 & 11.0 & 31.5 & 11.4 & 9.5 & 41.2 & 20.5 & 58.7 & 47.7  \\
\textbf{DAN} & 35.5 & 61.0 & 55.2 & 28.7 & 51.4 & 44.8 & 31.0 & 27.7 & 61.8 & 38.4 & 37.1 & 64.9 & 44.8 & 61.7 & 56.8 \\
\textbf{DANN} & 20.3 & 38.5 & 31.9 & 21.6 & 32.3 & 30.5 & 25.1 & 23.1 & 49.3 & 27.3 & 27.6 & 52.2 & 31.6 & 64.7 & 55.1 \\
\textbf{ADDA} & 19.2 & 39.7 & 33.6 & 17.9 & 27.5 & 28.5 & 22.3 & 18.6 & 46.3 & 29.0 & 23.5 & 53.2 & 29.9 & 65.3 & 55.1 \\
\textbf{PADA} & 10.7 & 32.3 & 17.8 & 16.1 & 24.0 & 24.8 & 17.1 & 17.8 & 44.8 & 17.6 & 16.9 & 45.1 & 23.7 & 65.4 & 53.4 \\
\textbf{MDD} & 13.5 & 29.9 & 25.3 & 16.1 & 26.1 & 28.7 & 18.5 & 19.1 & 42.4 & 20.5 & 21.6 & 48.5 & 25.9 & 65.5 & 54.0 \\
\textbf{SWD} & 17.6 & 38.1 & 23.7 & 20.2 & 27.2 & 27.2 & 22.6 & 20.1 & 45.6 & 20.5 & 20.0 & 49.5 & 27.7 & 64.8 & 54.1 \\
\textbf{CDAN} & 29.8 & 54.6 & 44.3 & 16.1 & 41.3 & 30.3 & 34.8 & 31.4 & 64.5 & 36.2 & 32.6 & 61.7 & 39.8 & 65.5 & 58.1 \\
\textbf{MCD-DA} & 19.2 & 41.7 & 27.5 & 23.7 & 32.4 & 33.9 & 26.0 & 21.8 & 49.8 & 27.3 & 24.5 & 50.2 & 31.5 & 64.6 & 55.0 \\
\textbf{DIFA} & 40.8 & 54.1 & 52.9 & 32.1 & 42.0 & 52.6 & 38.1 & 34.6 & 59.0 & 42.6 & 38.8 & 67.2 & 46.2 & 64.3 & 58.7 \\
\textbf{SAFN} & 18.7 & 47.3 & 30.5 & 21.9 & 39.9 & 29.2 & 21.8 & 20.4 & 46.5 & 27.9 & 26.1 & 49.5 & 31.6 & \textbf{65.8} & 56.0 \\
\hline
\textbf{GFCA-2Stage} & 41.0 & 66.0 & 64.5 & \textbf{38.7} & 58.6 & 58.2 & 42.6 & 36.1 & 66.8 & 46.1 & 44.6 & 69.9 & 52.8 & 64.7 & 61.3\\
\textbf{GFCA-WoFC} & 43.0 & 66.4 & 63.9 & 36.8 & \textbf{59.3} & 57.8 & \textbf{44.4} & \textbf{37.4} & 69.6 & \textbf{46.9} & 45.0 & 72.7 & 53.6 & 64.6 & 61.5  \\
\textbf{GFCA} & \textbf{43.2} & \textbf{67.5} & \textbf{65.6} & 37.5 & \textbf{59.3} & \textbf{60.2} & 43.9 & 36.5 & \textbf{70.5} & 46.1 & \textbf{45.4} & \textbf{74.5} & \textbf{54.2} & 64.7 & \textbf{61.7} \\
\hline
\end{tabular}
\vspace{-6mm}
\label{tbl:officehome}
\end{table*} 
\vspace{-2mm}
\subsection{Datasets}
\vspace{-2mm}
\textbf{Office31} \cite{gong2012geodesic} consists of images within 31 categories from 3 domains (Amazon, Webcan, and DSLR). Specifically, Amazon images are downloaded from online merchants with clear backgrounds. Webcam and DSLR images are taken from office environments, where Webcam images are taken by a low-resolution web camera and DSLR images are taken by a high-resolution digital SLR camera. 

\textbf{Office-Home} \cite{venkateswara2017deep} consists of images within 65 categories from 4 domains (Art, Clipart, Product, and Real-world). Specifically, Art represents artistic depictions of objects; Clipart contains clipart images; Product consists of images of objects without a background, similar to the Amazon domain in the Office31 dataset; Real-world contains images of objects captured with a regular camera.

\vspace{-2mm}
\subsection{Comparison Experiments}
\vspace{-2mm}
We compare our proposed GFCA algorithm with the following state-of-the-art DA methods: 
\textbf{DAN} \cite{long2015learning}, \textbf{DANN} \cite{ganin2014unsupervised}, \textbf{ADDA} \cite{tzeng2017adversarial}, \textbf{PADA} \cite{cao2018partial}, \textbf{MDD} \cite{zhang2019bridging}, \textbf{SWD} \cite{lee2019sliced}, \textbf{CDAN} \cite{long2018conditional}, \textbf{MCD-DA} \cite{saito2018maximum}, \textbf{DIFA} \cite{volpi2018adversarial}, and \textbf{SAFN} \cite{xu2019larger}. 

Specifically, DAN, SWD, MCD-DA, and SAFN are discrepancy-based methods, while DANN, ADDA, PADA, MDD, CDAN, and DIFA are adversarial-based methods. For baseline comparison, we also directly train a network with the same structure as the encoder and classifier in GFCA using only ResNet-50 \cite{he2016deep} features from the source domain (\textbf{ResNet-50}).

For the Office31 dataset, 10 classes are selected as few-shot classes. For the Office-Home dataset, 20 classes are selected as few-shot classes. Each few-shot class consists of 3 samples randomly sampled from the original dataset.
In all our experiments, source domain data are first randomly over-sampled to create a balanced training set. For fair comparisons across different methods, ResNet-50 \cite{he2016deep} pre-trained on the ImageNet dataset and fine-tuned using the source domain training samples is utilized as the backbone of the compared deep DA methods.
We adopt the top-1 classification accuracy for the unlabeled target samples as the evaluation metric. The results on the Office31 dataset and the Office-Home dataset are shown in Table \ref{tbl:office31} and Table \ref{tbl:officehome}, respectively. 

From Tables \ref{tbl:office31} and \ref{tbl:officehome}, we observe that all compared methods achieve similar classification accuracy on the normal set, which are much higher than the baseline. This indicates that DA is necessary for effective knowledge transfer across domains with different distributions. Specifically, SAFN obtains the highest normal set classification accuracy on the Office-Home dataset, which demonstrates the effectiveness of its novel approach of feature norm adaptation. However, the proposed GFCA algorithm performs the best in normal set classification on the Office31 dataset, which represents that effective data augmentation through generative models could also benefit the learning of normal classes.

For few-shot classification, we observe that the proposed GFCA algorithm makes remarkable performance improvement in all DA tasks comparing with existing DA methods on both datasets. This demonstrates that our generative few-shot cross-domain alignment approach could indeed boost few-shot classification performance and fair classification among normal and few-shot classes.
Existing methods generally perform better in few-shot classification when the size of the source dataset is small (e.g. D$\rightarrow$W and W$\rightarrow$D in Office31) since the imbalance between few-shot and normal set is less compelling. Surprisingly, the classical discrepancy-based DAN method achieves the best performance in few-shot classification of the Office31 dataset comparing with other existing methods. However, we also notice that DAN tends to obtain lower accuracy for normal set classification due to less cross-domain adaptation ability. 
DIFA achieves the best performance in few-shot classification of the Office-Home dataset comparing with other existing methods, which demonstrates that feature augmentation does provide help in effective domain adaptation for few-shot classes.
Under more difficult DA tasks in the Office-Home dataset, the classification performance improvement produced by GFCA becomes much more significant. This demonstrates the necessity of generative data augmentation under complex DA problems with extremely imbalanced source data.
Moreover, GFCA consistently achieves the highest overall classification accuracy comparing to existing DA approaches.

\vspace{-2mm}
\subsection{Empirical Evaluation}
\vspace{-2mm}
\begin{figure}[ht]
\centering
\begin{subfigure}[b]{0.65\linewidth}
    \includegraphics[width=\textwidth]{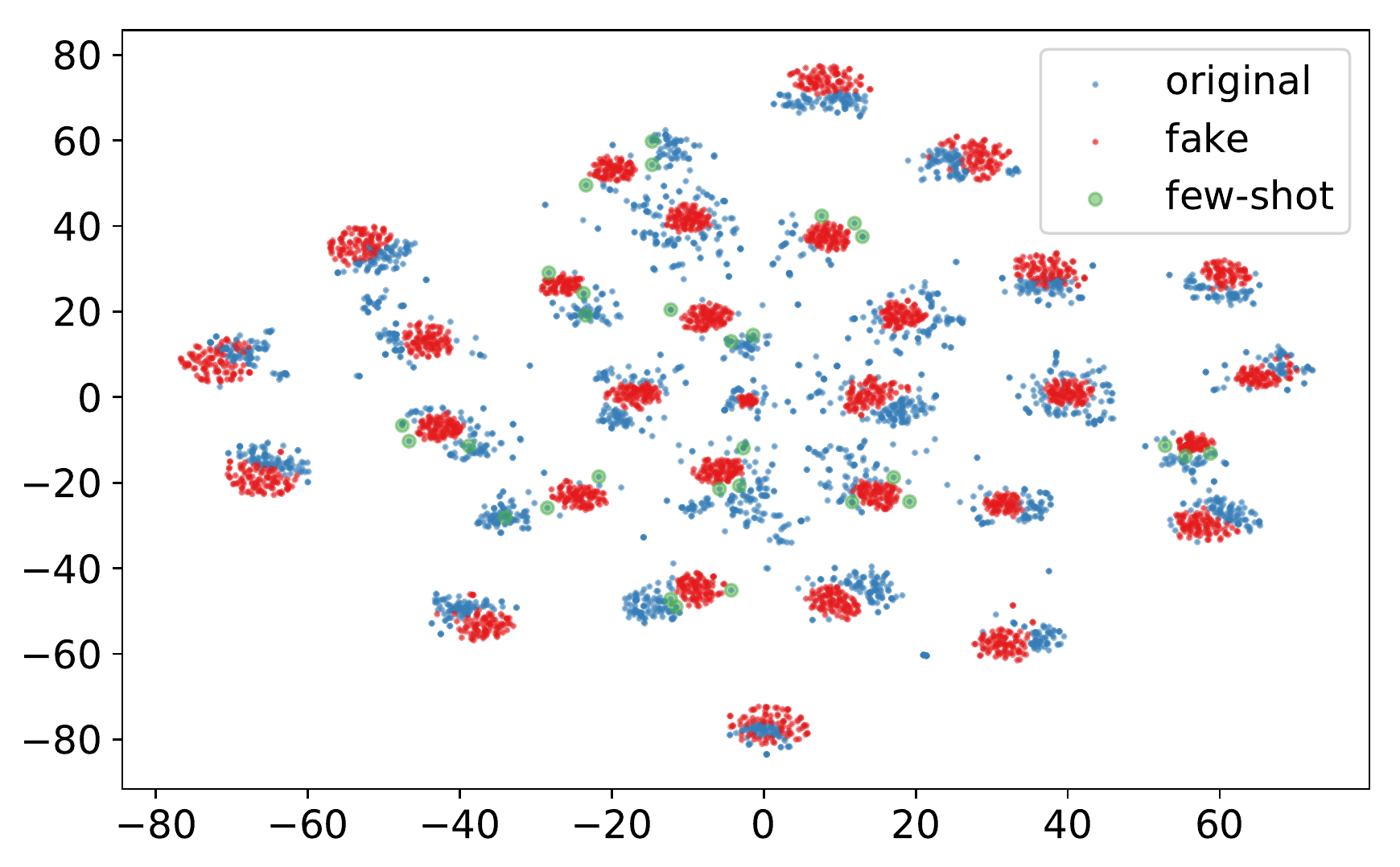}\vspace{-3mm}
    \caption{}
    \label{fig:office31_AD_tsne}
\end{subfigure}
\begin{subfigure}[b]{0.65\linewidth}
    \includegraphics[width=\textwidth]{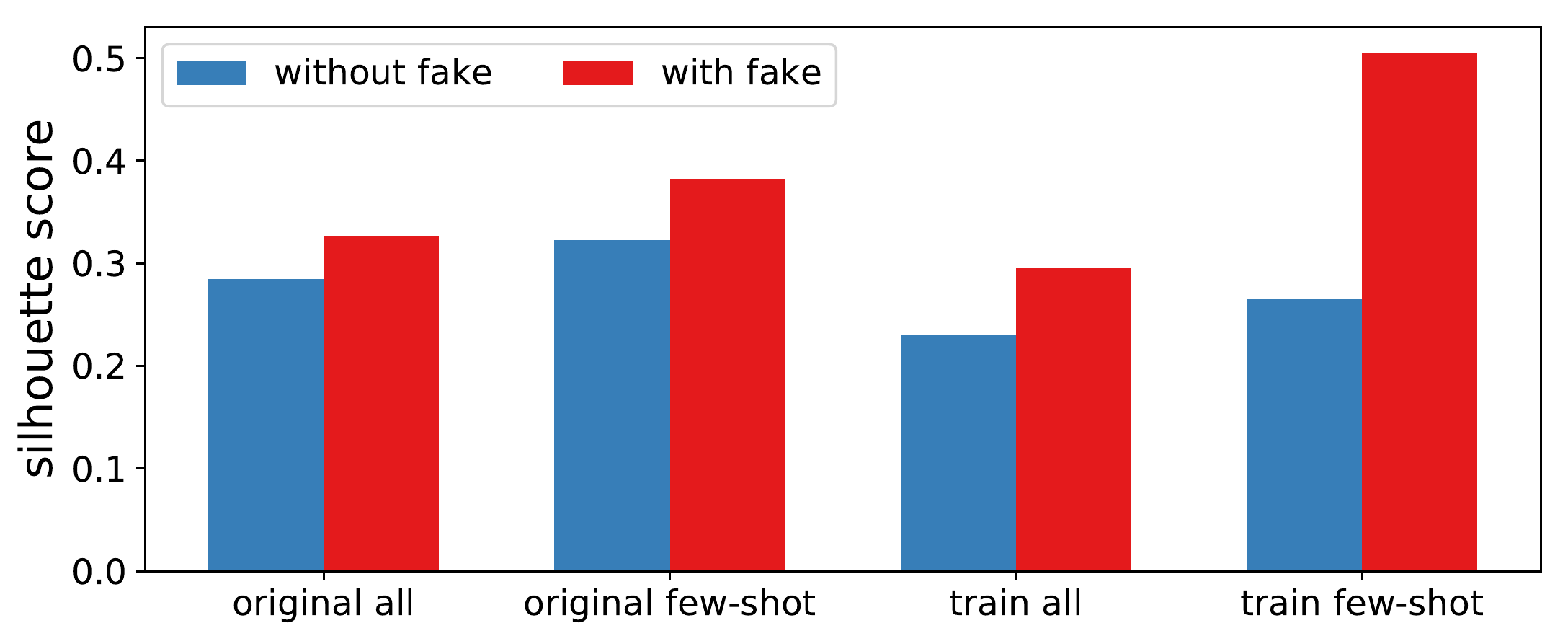}\vspace{-3mm}
    \caption{}
    \label{fig:office31_AD_sil}
\end{subfigure}
\begin{subfigure}[b]{0.65\linewidth}
    \includegraphics[width=\textwidth]{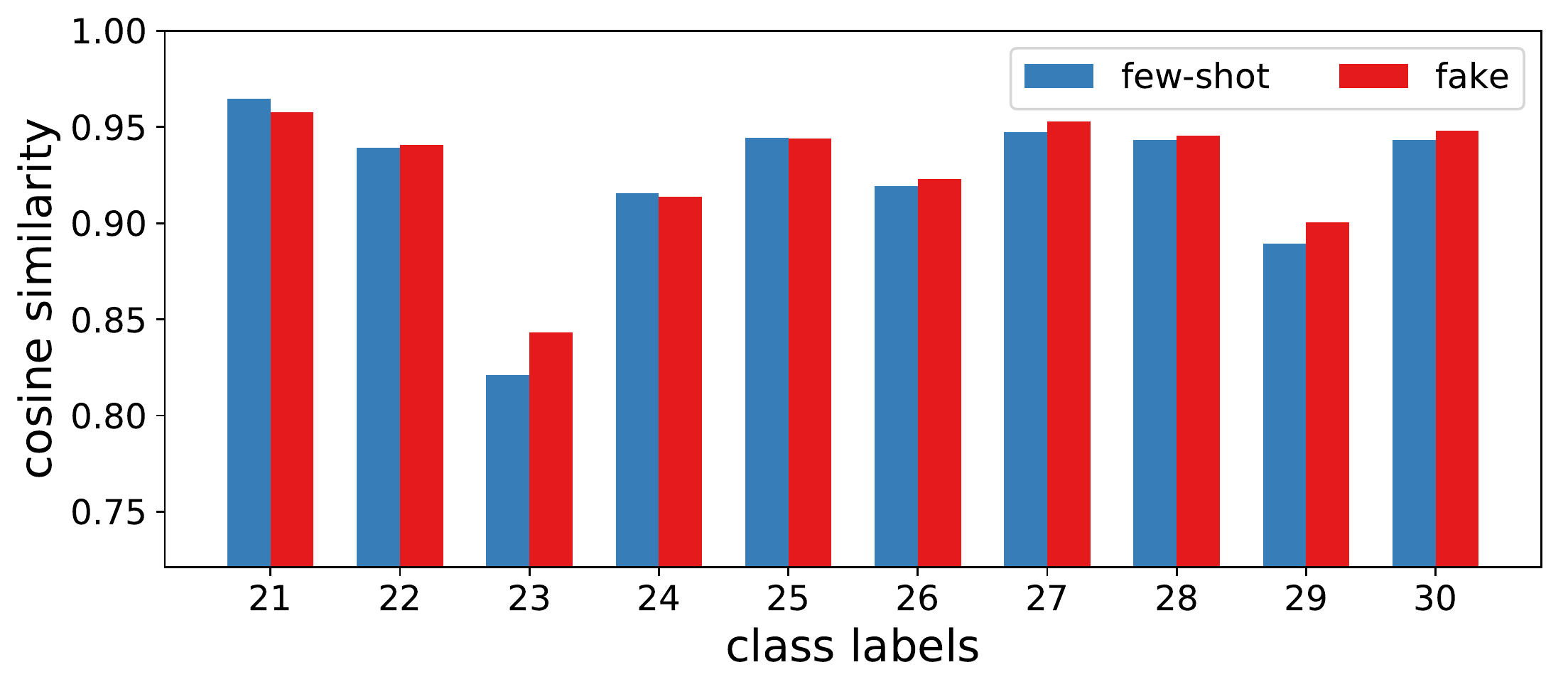}\vspace{-3mm}
    \caption{}
    \label{fig:office31_AD_center}
\end{subfigure} \vspace{-4mm}
\caption{Visualization of A$\rightarrow$D task in the Office31 dataset. (a) t-SNE visualization with synthetic fake data (red), sampled real few-shot source training data (green), and source data from the original full dataset (blue). (b) Silhouette scores comparing data with or without synthetic fake data with clusters defined by class labels. "All" represents silhouette scores calculated using all samples and "few-shot" represents silhouette scores calculated with few-shot samples. (c) Cosine similarity of class centers between few-shot source training data (blue) or synthetic fake source data (red) and the unused few-shot source data from the original full dataset.}
\label{fig:office31_AD}
\vspace{-8mm}
\end{figure}

To demonstrate that our generative data augmentation model could indeed synthesize effective training data and promote fair classification across both normal and few-shot set, we utilize t-SNE \cite{maaten2008visualizing} to visualize generated fake source data, along with source data in the training set and unused few-shot source data from the original full dataset. Figure \ref{fig:office31_AD_tsne} shows the few-shot t-SNE visualization for A$\rightarrow$D task in the Office31 dataset. We observe that synthetic few-shot data generally conform with the unused source samples in the full dataset for both few-shot and normal classes. Specifically, different classes form distinct clusters and the synthetic data generally spread around the center of each cluster. For few-shot classes, the generated data also expands the intra-class variation comparing with the few-shot training samples. We further evaluate the quality of the generated data using silhouette scores with clusters defined by class labels and distance calculated using cosine distance (Figure \ref{fig:office31_AD_sil}). We observe that when combined with the synthetic fake data, silhouette scores of the original full dataset and the training source data both increases. Such observation is also consistent when all samples or only few-shot samples are considered. This indicates that the synthetic data could enhance the clustering structure of the data for normal and few-shot classes, which could benefit the following classification task. Moreover, from Figure \ref{fig:office31_AD_tsne}, we also notice that in cases where the sampled few-shot training data is far from the class centers estimated by the original full datasets, GFCA is able to generate data closer to the class centers. To further confirm this observation, we evaluate the cosine similarity between class centers of generated few-shot data and unused few-shot data, and compare it with the cosine similarity between class centers of sampled few-shot training data and unused few-shot data (Figure \ref{fig:office31_AD_center}). We observe that for 7 out of 10 few-shot classes in A$\rightarrow$D task in the Office31 dataset, the class centers of the synthetic fake data obtain higher similarity with the few-shot data in the original full dataset, which demonstrates that the generator could acquire a better estimation of class centers through adversarial training. 

\begin{figure*}[ht]
\centering
\begin{subfigure}[b]{0.27\linewidth}
    \includegraphics[width=\textwidth]{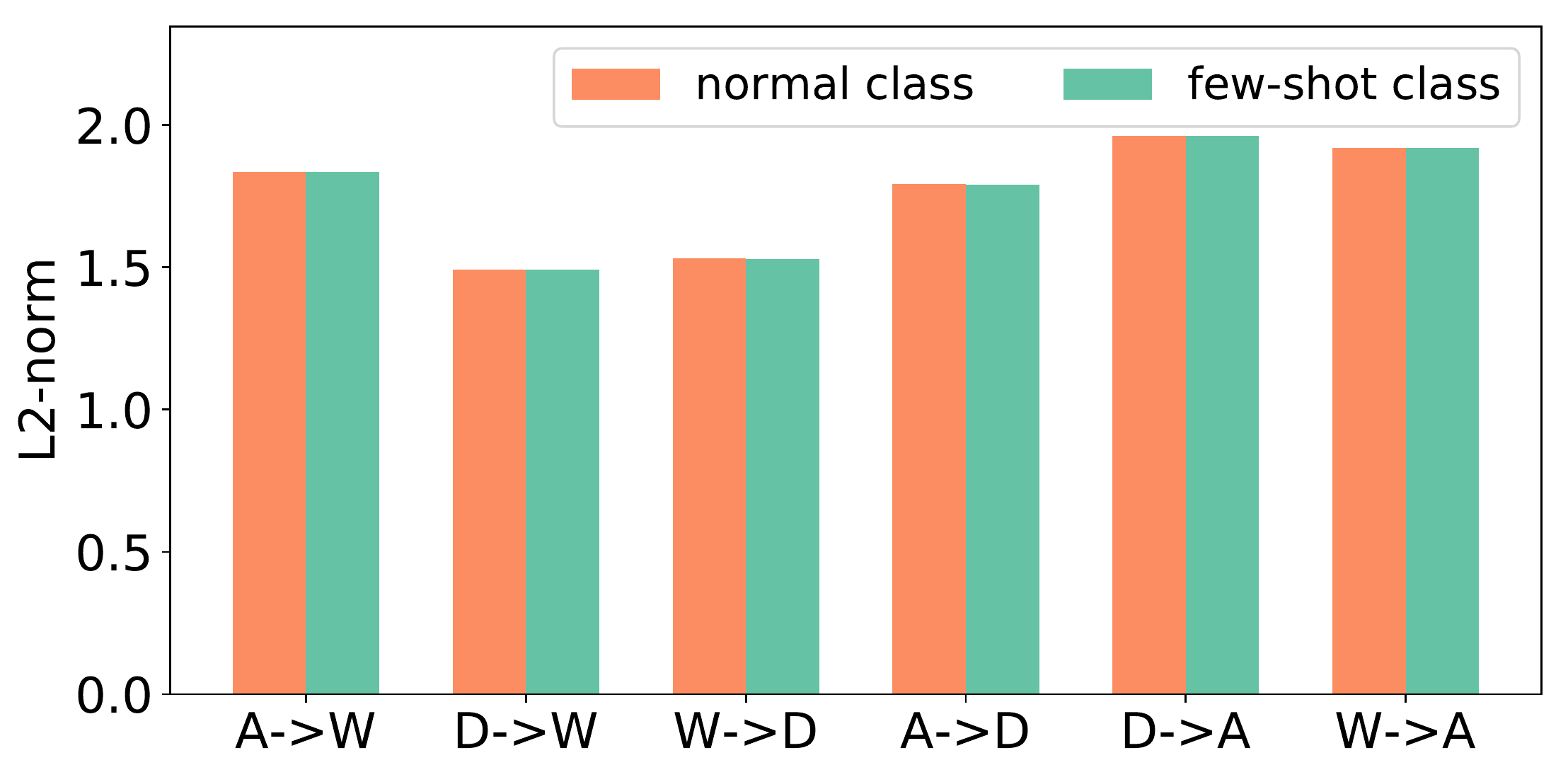}
    \caption{GFCA}
    \label{fig:clf_gfca}
\end{subfigure}
\begin{subfigure}[b]{0.27\linewidth}
    \includegraphics[width=\textwidth]{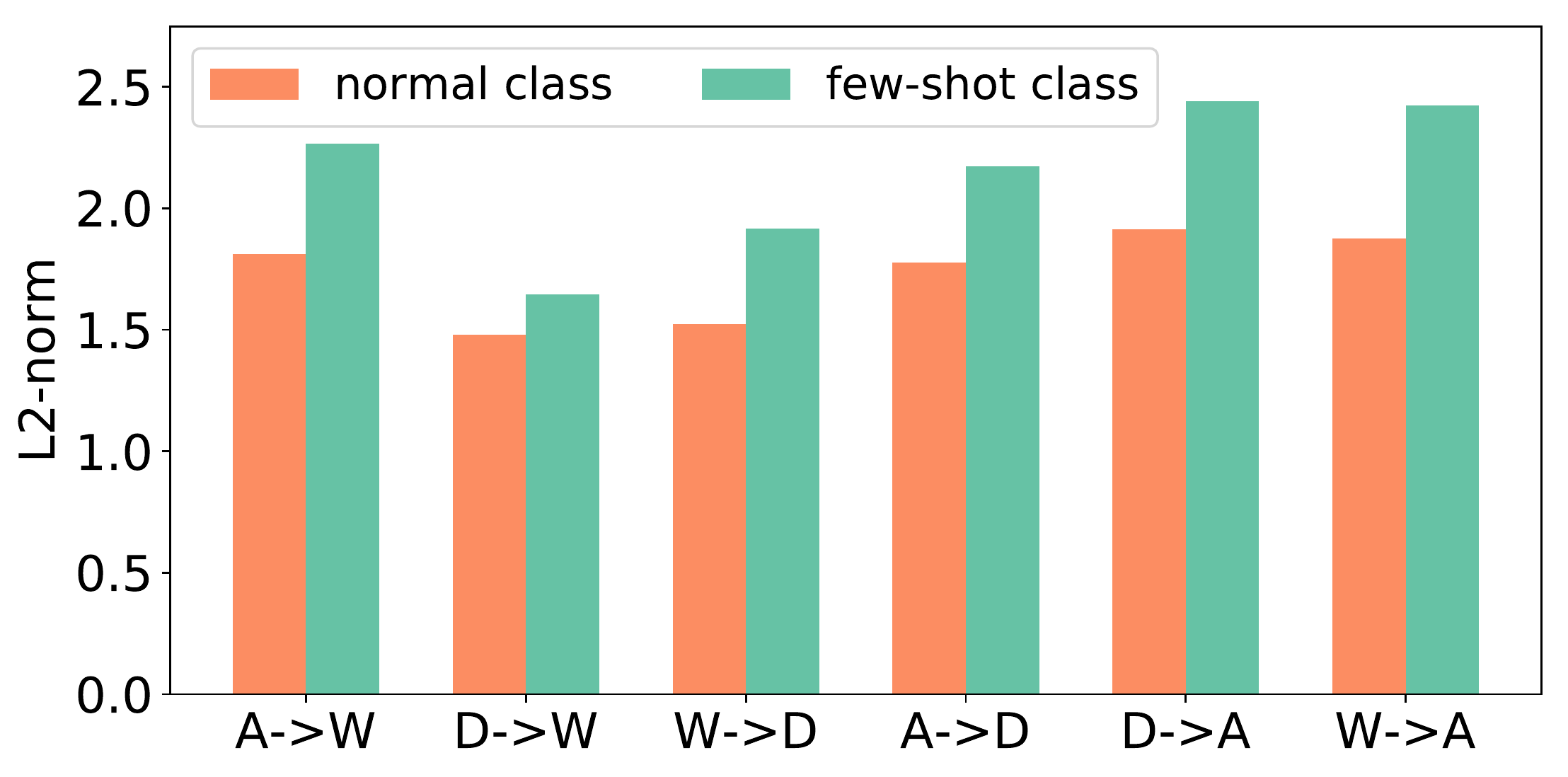}
    \caption{GFCA-WoFC}
    \label{fig:clf_wocb}
\end{subfigure}
\begin{subfigure}[b]{0.27\linewidth}
    \includegraphics[width=\textwidth]{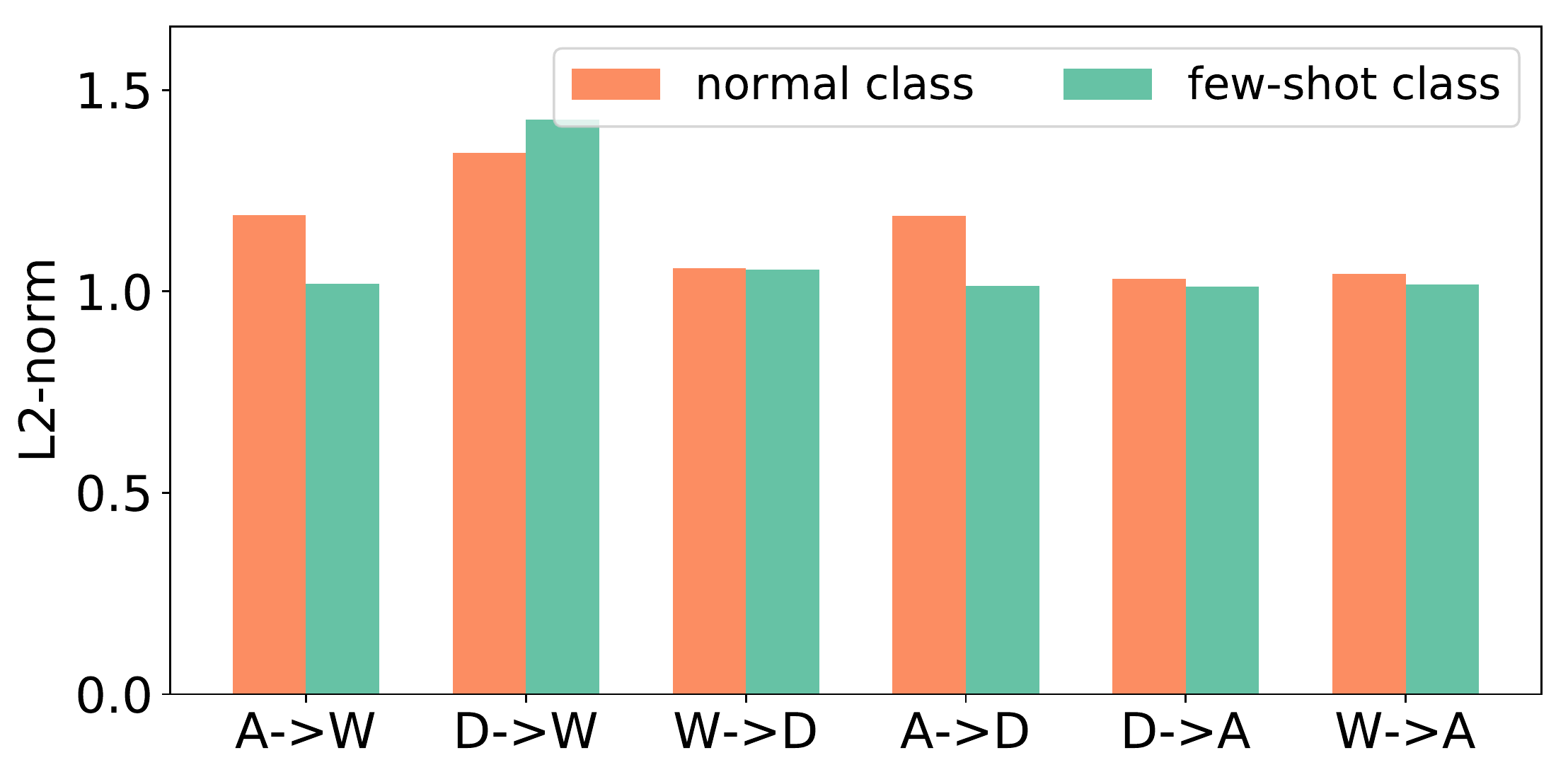}
    \caption{DANN}
    \label{fig:clf_dann}
\end{subfigure} \vspace{-4mm}
\caption{Average of L2-norms of the classifier weight vectors for normal classes (orange) and few shot classes (green) in different DA tasks of the Office31 dataset.}
\label{fig:clf}
\vspace{-6mm}
\end{figure*}

We further investigate the reason for few-shot classification improvement achieved by our method. The weight matrix $W_c$ in classifier $C(\cdot)$ consists of a weight vector $w_k$ for each class $k$. 
Guo et al. revealed that there exists a close connection between the norm of the weight vector and the volume of the corresponding class partition in the feature space \cite{guo2017one}.
Therefore, we calculate the norm of each class weight vector after training completed to evaluate whether the few-shot classes have been promoted with similar importance as the normal classes by the classifier. 
Figure \ref{fig:clf} shows the average L2-norms of normal and few-shot class weight vectors in GFCA and DANN under different DA tasks in the Office31 dataset. DANN is trained with the same classifier structure as GFCA for fair comparison. 
We notice for tasks with more imbalanced training set (e.g. A$\rightarrow$W, A$\rightarrow$D), weight norms of few-shot classes produced by DANN are significantly lower than norms of the normal set, which further verifies the performance deterioration of DANN under tasks with more imbalanced training data. 
By introducing the fair classification term, GFCA yields similar weight norms for both normal and few-shot sets, which conforms with the superior performance of GFCA in few-shot set, normal set, and overall classification.

\begin{figure}[ht]
\centering
\begin{subfigure}[b]{0.45\linewidth}
    \includegraphics[width=\textwidth]{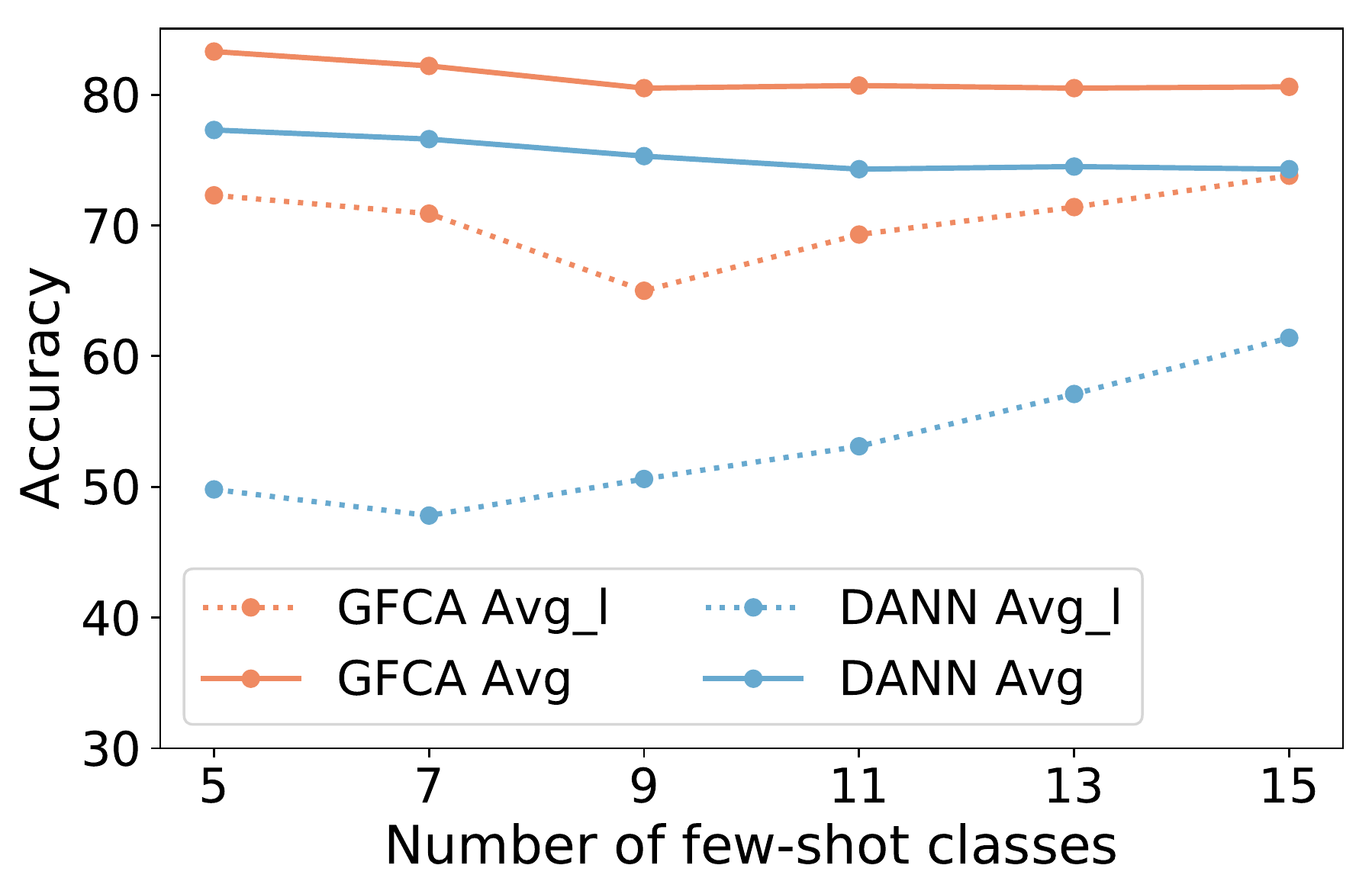}
    \caption{}
    \label{fig:office31_class}
\end{subfigure}
\begin{subfigure}[b]{0.45\linewidth}
    \includegraphics[width=\textwidth]{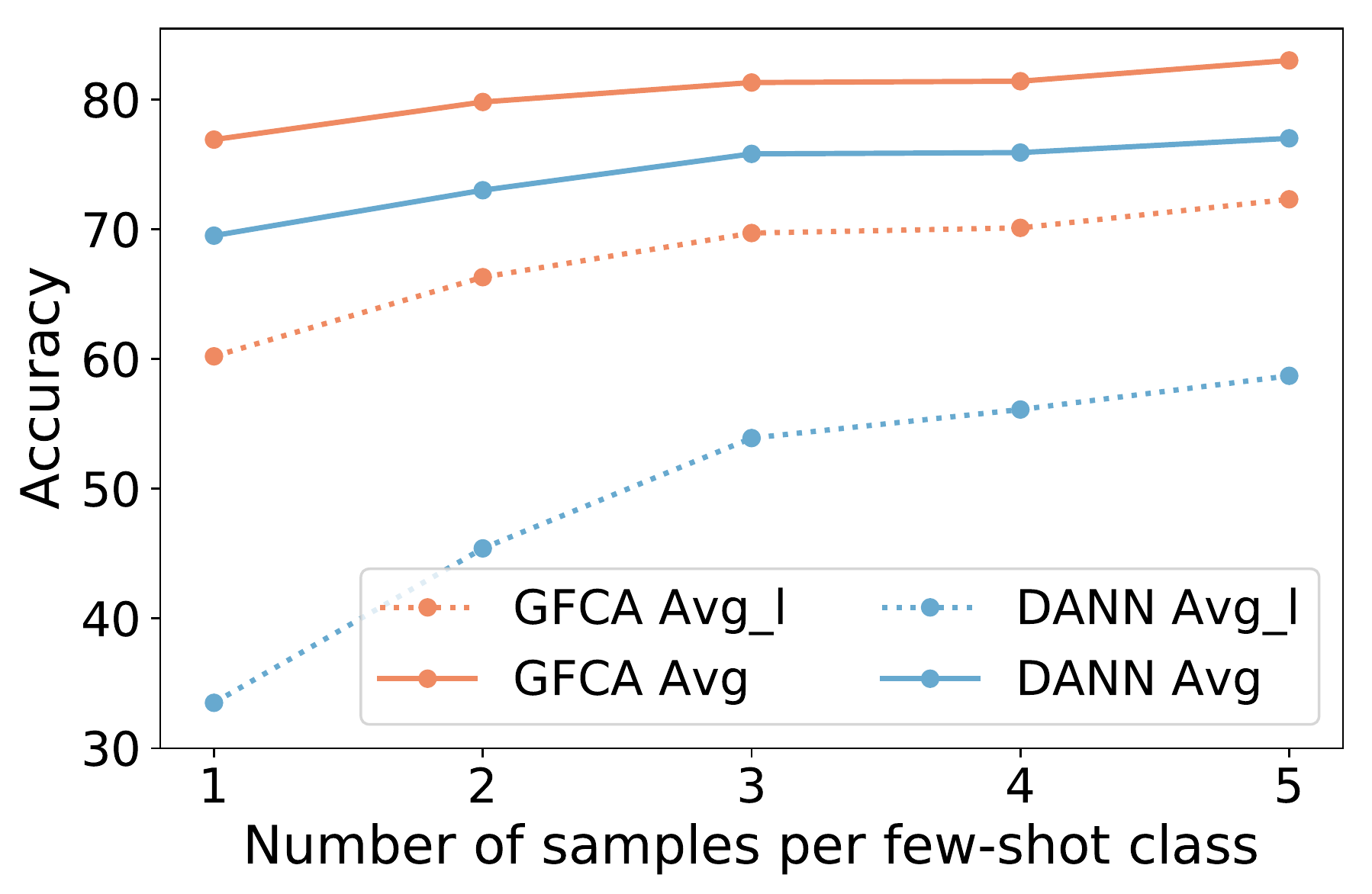}
    \caption{}
    \label{fig:office31_sample}
\end{subfigure}\vspace{-4mm}
\caption{Few-shot and overall classification accuracy of GFCA and DANN under different few-shot settings in the Office31 dataset. (a) Classification accuracy with respect to the number of few-shot classes. (b) Classification accuracy with respect to the number of samples per few-shot class.}
\label{fig:class_sample}
\vspace{-4mm}
\end{figure}

We also evaluate the performance of GFCA under different few-shot settings. Specifically, two factors are investigated: number of few shot classes and number of samples per few-shot class. Firstly, we fix the number of samples per few-shot class to 3 and change the number of few-shot classes from 5 to 15 (Figure \ref{fig:office31_class}). Secondly, we fix the number few-shot class to 10 and change the number of samples per few-shot classes from 1 to 5 (Figure \ref{fig:office31_sample}). 
For meaningful and consistent comparison, we choose the same set of few-shot samples for the same few-shot class across different experiments when exploring the influence of a factor. For example, in the experiments with 5 and 9 few-shot classes, the shared 5 few-shot classes across these two experiments contain the same set of samples. The few-shot cross-domain learning problem becomes more difficult as the number of few-shot class increases or the number of samples per few-shot class decreases.
We also train DANN using the same data for comparison.
From Figure \ref{fig:class_sample}, we observe that GFCA consistently achieves remarkable improvement in both few-shot and overall classification under all few-shot settings. 
Specifically, when the number of few-shot classes increases, the overall classification accuracy in both methods gradually decreases. However, we observe fluctuation in few-shot classification accuracy under different number of few-shot classes, which could relate to the random sampling of few-shot training samples, as well as the variation in difficulty of classifying different categories. For DANN, the decrease in the number of normal classes potentially makes the model more difficult to learn the normal set while probably more generalizable to the few-shot set.
When the number of samples per few-shot class increases, we observe that the performance of both methods increases, since larger samples per few-shot classes represents less imbalanced dataset and easier few-shot cross-domain learning problem.
\vspace{-3mm}

\vspace{-2mm}
\subsection{Ablation Study}
\vspace{-2mm}
We compare GFCA with its two variants. The first variant is \textbf{GFCA-2Stage}, where data augmentation and domain adaptation in GFCA are performed in two stages. In GFCA-2Stage, the generator and discriminator are first trained using the labeled source domain data. Then, the generator is utilized to synthesize fake data for source domain data augmentation using balanced class labels. Finally, the encoder and classifier are trained using the augmented source data and target data. The second variant is \textbf{GFCA-WoFC}, where the FC regularization term on the classifier is removed.

From Table \ref{tbl:office31} and \ref{tbl:officehome}, we observe that all variants of the GFCA algorithm achieve significant improvement in few-shot and overall classification under all DA tasks comparing to existing methods. Specifically, on average, GFCA performs better in few-shot and overall classification than GFCA-2Stage, which demonstrates the advantage of end-to-end training of our proposed model. 
GFCA-WoFC yields the highest accuracy for few-shot classification in the Office31 dataset. However, it also obtains lower normal set and overall classification accuracy comparing to GFCA.
From Figure \ref{fig:clf_wocb}, we observe that classifiers in GFCA-WoFC have larger norms of few-shot class weight vectors than norms of normal class weight vectors. This indicates that generative data augmentation can synthesize data with enough intra-class variation and expand the volume of few-shot classes in the feature space. 
Although generative data augmentation alone could significantly improves few-shot classification, regularization with FC term is still needed to maintain the performance in normal set classification and the classification fairness across both few-shot and normal classes. 
Moreover, in the Office-Home dataset, we observe that GFCA achieves higher accuracy in both few-shot and overall classification comparing to GFCA-WoFC, which demonstrates that FC term is still needed for few-shot promotion and fair classification under more difficult DA tasks.

\vspace{-2mm}
\section{Conclusion}
\vspace{-2mm}
In this paper, we propose a novel GFCA algorithm to solve the few-shot cross-domain adaptation problem for fair classification. We utilize generative data augmentation to synthesize effective source training data for few-shot classes and alleviate the bias in favor of the normal classes in the training set, where we train a conditional generative adversarial network to capture the intra-class variation of the normal classes. We then leverage effective domain alignment to adapt knowledge from the source domain to the target domain, where MMD-based regularization is utilized to learn transferable representations of training samples using real source data, fake synthetic source data, and target data. A general classifier is further trained with a FC regularization term to balance between the underrepresented few-shot classes and normal classes for fair classification. Experiments on two cross-domain benchmark datasets demonstrate that our method could significantly improve the performance for both few-shot and overall classification comparing to state-of-the-art DA methods. 

\clearpage

{\small
\bibliographystyle{ieee_fullname}
\bibliography{ref}
}

\end{document}